\documentclass[runningheads]{llncs}

\usepackage[T1]{fontenc}
\usepackage{amsmath,amssymb,amsfonts}
\usepackage{graphicx}
\usepackage[expansion=false]{microtype}
\usepackage[hidelinks]{hyperref}

\newcommand{\code}[1]{\texttt{\small #1}}

\begin{document}

\title{Knowledge Graphs and Explainable AI as Complementary Resources for Urban Mining}
\titlerunning{KG--XAI Complementarity for Pre-Demolition Assessment}

\author{Jan Gronewald \and Andreas Emrich \and Nijat Mehdiyev}
\authorrunning{J.~Gronewald et al.}

\institute{German Research Center for Artificial Intelligence (DFKI), Saarbr\"ucken, Germany\\
\email{\{jan.gronewald,andreas.emrich,nijat.mehdiyev\}@dfki.de}}

\maketitle

\begin{abstract}
Pre-demolition assessment, the regulated audit process at the heart of urban mining, is an information process in which AI support must serve qualified auditors who remain accountable for the decisions taken. The relevant unit of value is not prediction accuracy alone, but the \emph{defensibility} of the supported decisions: their legibility, plausibility, sourcing, and contestability. Explainable AI techniques and domain knowledge graphs each address parts of this requirement, and existing taxonomies have catalogued their integration. The literature is descriptively rich but structurally under-specified: what remains less developed is a structural account of why specific integrations produce artefacts neither resource can provide alone. This paper offers a complementarity-theoretic interpretation grounded in the IS resource-based tradition. We propose four consolidated KG--XAI integration modes (Lifting, Constraining, Typing, and Revising), each defined as a typed operation over XAI artefacts and knowledge-graph substrate structures. Each mode unlocks a distinct property of defensibility and contributes to the kind of regulatory artefact pre-demolition assessment demands. A fire-door example from the urban-mining process illustrates the modes using the W3C Linked Building Data stack and valuation extensions.

\end{abstract}

\section{Introduction}

Urban mining, the systematic recovery of secondary materials from existing building stock, depends on a regulated audit process at the building's end of life: pre-demolition assessment (PDA). PDA structures this process as a multi-stage workflow in which a qualified auditor inventories components, characterises their materials and conditions, and routes them toward reuse, recycling, or disposal; in Germany the workflow is formalised by DIN SPEC 91484:2023~\cite{din2023}. The performance of urban mining at scale is bounded less by the accuracy of any single technical component embedded in this process than by whether the decisions it supports are \emph{defensible}: explainable to the auditor, traceable in documentation, and accountable to the regulatory regime under which they are taken~\cite{honic2019}.

Two technological resources speak to this requirement. Explainable AI offers a now-standard repertoire of techniques, such as feature attribution~\cite{ribeiro2016,lundberg2017} and counterfactual explanation~\cite{wachter2017,mothilal2020}, backed by a social-science account of useful explanations~\cite{miller2019}. These techniques operate on the input space of the underlying model, however, and their outputs are correspondingly expressed in that space rather than in the regulatory categories within which the auditor must document. Symbolic structures are therefore needed to translate model-facing explanations into audit-facing categories. Knowledge graphs allow heterogeneous domain knowledge to be defined, linked, queried, and reasoned over at both conceptual and instance levels. The W3C Linked Building Data Working Group has defined a set of matching ontologies for buildings (semantic descriptions of elements, floor plans, material compositions, geometries, and product and pricing data~\cite{rasmussen2021}), with valuation indices such as the Urban Mining Index~\cite{rosen2021} providing the circular-construction layer on top.

The integration of these resources has been taxonomised before~\cite{lecue2020,tiddi2022}. The existing literature is descriptively rich but structurally under-specified. Where existing taxonomies classify KG--XAI integration by direction, function, or technical mechanism, we classify by the \emph{defensibility property} produced by the joint KG--XAI object: the contribution is not a new inventory of techniques but a structural account of how technical pairings become accountability-bearing audit artefacts.

This paper makes three contributions. It provides a complementarity-theoretic interpretation of KG--XAI integration for accountability-bearing expert processes, drawing on the IS resource-based tradition~\cite{bharadwaj2000,melville2004}. It consolidates prior KG--XAI integration patterns into four typed modes (Lifting, Constraining, Typing, and Revising), each linked to a distinct defensibility property. It illustrates the framework through a fire-door example grounded in the W3C Linked Building Data stack and the Urban Mining Index; the paper is positional rather than empirical.

\section{Technology Complementarity as Methodological Lens}

Two resources are complementary, in the IS resource-based view, when the marginal contribution of one increases in the level of the other~\cite{bharadwaj2000,melville2004}, an account that has explained disappointing standalone returns of IT investments absent complementary process maturity and trust-shaping mechanisms~\cite{shneiderman2022}. Applied to KG--XAI, the claim is structural rather than aggregative: the joint object of an integrated operation has properties that the unilateral object lacks: \emph{legibility} in the regulatory vocabulary, \emph{plausibility} under domain constraints, \emph{sourcing} of uncertainty, and \emph{contestability} against named claims. An XAI counterfactual without ontology-based constraints may still be an explanation, but it need not be plausible in the physical or regulatory sense. An ontology fragment without XAI is still a knowledge artefact but cannot be responsive to a particular instance. Each resource unlocks a distinct property the other cannot supply, and the four properties together constitute the defensibility profile targeted here for urban-mining audit artefacts.

The framework operates on the explanatory surface of black-box predictors and does not extend without modification to inherently interpretable models~\cite{rudin2019}. Trustworthiness moderators (explanation, uncertainty, contestability) enter as properties of joint objects rather than as parallel features~\cite{shneiderman2022,jacovi2021}. Explanations alone can raise users' confidence without raising their accuracy when they lack the capacity to scrutinise them~\cite{lai2019}; the structural argument is that the symbolic substrate is what allows the user's accountability to engage with the explanatory surface at all.

\section{Four Modes of KG--XAI Complementarity}\label{sec:modes}

The four modes operate on a common object: the property-state graph of an instance. Let $\mathcal{X}$ be the input space of the predictor $f : \mathcal{X} \to \mathcal{Y}$, and let $\mathcal{K}$ be the knowledge graph with controlled vocabulary $\mathcal{V}_{\mathcal{K}}$ and instance set $\mathrm{Inst}(\mathcal{K})$. A \emph{property assertion} about an instance $x$ is a tuple $(e, p, v, r)$ with $e \in \mathrm{Inst}(\mathcal{K})$, $p \in \mathcal{V}_{\mathcal{K}}$, $v$ a value, and $r \in \mathcal{R}$ a reliability annotation drawn from the substrate's typology; for the LBD stack, $\mathcal{R} = \{\mathrm{Assumed}, \mathrm{Confirmed}, \mathrm{Deleted}\}$ paired with a numeric score. A \emph{property-state graph} is a set $\Sigma(x)$ of such assertions. We treat the formalism as notational scaffolding rather than operational semantics. Table~\ref{tab:modes} summarises the four modes; the paragraphs that follow develop each.

\begin{table}[t]
\caption{Four KG--XAI complementarity modes and their defensibility properties.}\label{tab:modes}
\small
\setlength{\tabcolsep}{4pt}
\begin{tabular}{@{}lp{0.18\textwidth}p{0.22\textwidth}p{0.22\textwidth}l@{}}
\hline
\textbf{Mode} & \textbf{XAI} & \textbf{KG} & \textbf{Joint product} & \textbf{Property} \\
\hline
Lifting & Attributions & Grounding, mereology & Property graph & Legibility \\
Constraining & Counterfactuals & Compatibility axioms & Plausible interventions & Plausibility \\
Typing & Evidence & Reliability types & Sourced assertions & Sourcing \\
Revising & Explanations & Provenance and state & Revision history & Contestability \\
\hline
\end{tabular}
\end{table}

\textbf{Mode 1 (Lifting)} populates $\Sigma(x)$ from XAI attributions. The XAI side supplies an attribution function $\alpha : \mathcal{X} \to \mathbb{R}^{|\mathrm{features}|}$ from methods such as SHAP or gradient attribution~\cite{ribeiro2016,lundberg2017}. The KG side supplies a \emph{grounding} $\gamma : \mathrm{features}(\mathcal{X}) \to \mathrm{Inst}(\mathcal{K})$ that maps each feature region to a typed instance (a salient region becomes a \code{product:DoorSeal} individual, a detected component a \code{product:DoorLeaf}), together with a mereological hierarchy $\mathcal{H}_{\mathrm{part}}$ of part-whole relations. Lifting has signature $\mathsf{Lift}_{\gamma, \mathcal{H}_{\mathrm{part}}} : (x, \alpha) \mapsto \Sigma(x)$: each input feature is grounded by $\gamma$ into a typed instance, salience masses are aggregated along $\mathcal{H}_{\mathrm{part}}$ from child components to parent components, and the result is a set of attribution-weighted property assertions about $x$ in $\mathcal{V}_{\mathcal{K}}$. Aggregation is restricted to mereology, since taxonomic refinement is not aggregable by sum and is the subject of Mode~3. The property unlocked is \emph{legibility}: the explanation lives in the regulatory vocabulary at the granularity at which it must be documented~\cite{miller2019,rosen2021}. Lifting requires $\gamma$ to cover the salient regions; under-coverage leaves high attribution mass ungrounded.

\textbf{Mode 2 (Constraining)} filters XAI counterfactual candidates through an intervention-typed plausibility predicate. The XAI side produces a candidate set $\mathcal{C}(x, f) \subseteq \mathcal{X}$ via methods such as Wachter-style optimisation or DiCE~\cite{wachter2017,mothilal2020}. Each candidate corresponds to an \emph{intervention} $\iota$ that names which property of which instance is being substituted. The KG side supplies, per intervention type and per entity type, the relevant compatibility axioms drawn from the substrate's axiomatisation: definitional dependencies, structural compatibility, recovery-pathway admissibility. Constraining has signature $\mathsf{Constr}_{\Phi_{\mathcal{K}}} : \mathcal{C}(x, f) \mapsto \mathcal{C}^{\mathrm{plaus}}(x, f)$, where $\Phi_{\mathcal{K}}(x' \mid \iota, e)$ is the conjunction of axioms in $\mathrm{Compat}(\iota, \mathrm{type}(e))$ that constrain the substitution $\iota$ for entities of type $\mathrm{type}(e)$. Crucially, $\Phi_{\mathcal{K}}$ is parameterised by the intervention rather than by all axioms in $\mathcal{K}$: a counterfactual must satisfy axioms constraining the substitution it proposes, not those it leaves unchanged. The output is a counterfactual surface alongside $\Sigma(x)$: surviving candidates name interventions the auditor could perform. The property unlocked is \emph{plausibility}: counterfactuals correspond to interventions admissible as a basis for a real audit decision. Selectivity depends on $\mathrm{Compat}$'s coverage in computable form; under-axiomatised substrates yield permissive filters.

\textbf{Mode 3 (Typing)} constructs reliability annotations on $\Sigma(x)$ from per-property evidence. The XAI side supplies, for each assertion $(e, p, v) \in \Sigma(x)$, an evidence chain: the predictor's confidence on that specific output, the chain of inferences in $\mathcal{K}$ that produced it, and calibration data from prior auditor confirmations. The KG side supplies the reliability typology $\mathcal{R}$ and an evidence-to-reliability operator $\Delta_{\mathcal{K}}$. Typing has signature $\mathsf{Type}_{\Delta_{\mathcal{K}}} : \Sigma_{\bot}(x) \mapsto \Sigma(x)$, taking a property-state graph with default reliability annotations and returning one whose annotations are typed and quantified per property. This is deliberately a construction from per-property evidence, not the decomposition of a single scalar, an operation not well-defined in the general case. The property unlocked is \emph{sourcing}: the auditor's response to each property is differentiated by the evidence supporting it. Typing requires that each assertion have a traceable evidence chain; assertions whose chains are missing receive a default reliability and become opaque to differentiated next-step action.

\textbf{Mode 4 (Revising)} is bidirectional, with within-instance and cross-instance operators. Within-instance, an auditor action (accept, contest a claim, or add a claim) feeds a transition operator $\tau_{\mathrm{inst}}$ that updates $\Sigma(x)$ and emits a PROV-O attribution chain naming the auditor, time, and action as derivation source~\cite{lebo2013}. Cross-instance, actions that contest an axiom or propose a refinement aggregate across instances into an evidence pool $E_{\mathcal{K}}$, feeding an ontology-revision process operating on $\mathcal{K}$ itself, distinct in timescale and authority. The within-instance operator runs at audit time under the auditor's individual accountability; the cross-instance operator runs at curation time under whatever governance regime the substrate carries. The KG side turns contestation into a recorded, queryable, longitudinal operation. The property unlocked is \emph{contestability}: the audit decision is not a single act but an inspectable, addressable, revisable artefact whose revision history is queryable~\cite{miller2019,lai2019}. Mode~4's most demanding precondition is the governance regime distinguishing acceptable revision from drift; OPM and PROV-O provide the \emph{mechanism} but not the \emph{authority} under which it occurs.

The four modes compose as a typed pipeline. Lifting and Typing produce the property-state graph, $\Sigma(x) = \mathsf{Type}_{\Delta_{\mathcal{K}}}(\mathsf{Lift}_{\gamma, \mathcal{H}_{\mathrm{part}}}(x, \alpha))$; Constraining produces the counterfactual surface, $\mathcal{C}^{\mathrm{plaus}}(x, f) = \mathsf{Constr}_{\Phi_{\mathcal{K}}}(\mathcal{C}(x, f))$; $\tau_{\mathrm{inst}}$ updates $\Sigma(x)$ in response to within-instance auditor action; $E_{\mathcal{K}}$ accumulates cross-instance evidence. The four modes are jointly necessary for the four-property defensibility profile targeted in PDA: removing Lifting empties $\Sigma(x)$; removing Typing leaves it with default reliability; removing Constraining leaves no plausible counterfactual surface; removing Revising leaves $\Sigma(x)$ static and produces no provenance trail. PDA workflows that do not target all four properties require correspondingly fewer modes; the framework characterises the available modes rather than mandating them.

\section{Worked Example: A Fire Door in Pre-Demolition Assessment}\label{sec:example}

The substrate is the W3C LBD stack: BOT for typed components and topology, PRODUCT for typed building products, PROPS for measured attributes, OPM for the property lifecycle~\cite{rasmussen2021,lebo2013,holten2018}, extended by the Urban Mining Index for recovery pathways~\cite{rosen2021}. The instance is a wooden door encountered during a Stage-1 PDA walkthrough of an office building; the detector returns \emph{door, 0.91}, a fine-grained classifier \emph{fire door, 0.62}.

\emph{Lifting} maps the XAI region attribution (three salient regions: seal between leaf and frame, leaf surface profile, hardware) to PRODUCT-typed instances (door seal, door leaf, door hardware), with PROPS-encoded fire-resistance class and material drawn into the explanation. The auditor receives an explanation that the classification rests on the seal, leaf profile, and hardware, all consistent with the fire-door pattern, and copies this directly into the audit's component-description field.

\emph{Constraining} filters the auditor's contrastive question (\emph{why is the predicted reuse value low?}) through the substrate's axioms. Counterfactuals perturbing the leaf-core composition to non-fire-rated material are rejected: the door is in a fire-compartment wall, an alignment encoded in BOT topology and propagated through PROPS-encoded fire-class compatibility. The counterfactual converting the connection from glued to bolted survives, yielding the actionable contrast: bolted rather than glued hinges would shift the recovery pathway from recycling to direct reuse and raise the predicted reuse value.

\emph{Typing} constructs reliability through OPM. The detector's evidence yields the door class as $\mathrm{Assumed}\,(0.91)$; the fine-grained classifier yields the fire-resistance class as $\mathrm{Assumed}\,(0.68)$; the substrate's recovery-pathway inference yields the recovery pathway as $\mathrm{Assumed}\,(0.94)$. The auditor's response is correspondingly typed: the moderate fire-resistance reliability triggers a clarifying inspection of the seal rather than a re-photograph or a referral.

\emph{Revising} closes the loop. On inspection, the seal is not intumescent but smoke-only; the door is a smoke-control door, a sub-class incompletely specified in the current substrate. The auditor contests the typical-of inference \emph{fire-door hardware} $\to$ \emph{fire door} by pointing to the PRODUCT/PROPS edge that supplied it. Through OPM, the fire-resistance class transitions from $\mathrm{Assumed}\,(0.68)$ to $\mathrm{Deleted}$ and a smoke-control class is $\mathrm{Confirmed}$, with PROV-O attribution naming auditor, time, and reasoning. Longitudinally, the contestation enters $E_{\mathcal{K}}$ as evidence for a refined typical-of relation distinguishing intumescent from smoke-only seals, a proposed rather than immediate substrate revision. The audit record retains the OPM transition chain in queryable form: the decision is corrected \emph{and} inspectable.

The example is illustrative rather than typical: many audits will exercise Lifting and Typing alone, without invoking Mode~2 or producing a Mode~4 contestation. Removing Lifting nonetheless leaves a saliency heat-map and no documentation; removing Constraining admits regulatorily forbidden substitutions when contrastive questions arise; removing Typing leaves an unsourced 0.62; removing Revising leaves a wrong inventory entry, no provenance trail, and a substrate that will repeat the error.

\section{Discussion and Conclusion}

The re-reading suggests that cautionary findings about XAI in expert-domain settings~\cite{rudin2019,lai2019} acquire a structural reading: explanations fail not because techniques are inadequate but because they are deployed without a substrate that unlocks the properties defensibility requires. The four-mode structure thereby identifies a route by which AI deployments in accountability-bearing processes can be made defensible by construction~\cite{shneiderman2022,jacovi2021}, bearing on regulatory regimes, such as the EU AI Act and sector-specific standards for the built environment, that increasingly demand documented reasons for automated decisions. The framework does not solve ontology governance: claims, state transitions, and proposed refinements must be reviewed under an institutional authority external to the substrate; OPM and PROV-O track such review but do not supply the authority. The framework is offered as a vocabulary for KG--XAI integration in expert-domain AI where accountability resides with a qualified human and the unit of value is decision defensibility. Future work will operationalise the framework in PDA case studies, evaluating whether KG-grounded explanations improve auditors' ability to document, contest, and revise AI-supported assessments.

\begin{credits}
\subsubsection{\ackname} This work was conducted within the MIRAKEL project, funded by the German Federal Ministry of Research, Technology and Space (BMFTR) under the FONA strategy ``Resource-efficient Circular Economy -- Urban Mining'' (grant number 033R422E).

\subsubsection{\discintname} The authors have no competing interests to declare that are relevant to the content of this article.
\end{credits}



\begin{thebibliography}{99}
\setlength{\itemsep}{0pt}
\setlength{\parskip}{0pt}

\bibitem{din2023}
German Institute for Standardization: DIN SPEC 91484:2023-09 - Procedure to record building materials as a base to evaluate the potential for a high-quality reutilization prior to demolition and renovation work (pre-demolition audit). DIN Media, Berlin, Germany (2023)

\bibitem{honic2019}
Honic, M., Kovacic, I., Rechberger, H.: Improving the recycling potential of buildings through Material Passports (MP): an Austrian case study. Journal of Cleaner Production \textbf{217}, 787--797 (2019)

\bibitem{ribeiro2016}
Ribeiro, M.T., Singh, S., Guestrin, C.: ``Why should I trust you?'': Explaining the predictions of any classifier. In: Proceedings of the 22nd ACM SIGKDD international conference on knowledge discovery and data mining, pp.~1135--1144. ACM, New York, NY, USA (2016)

\bibitem{lundberg2017}
Lundberg, S.M., Lee, S.-I.: A unified approach to interpreting model predictions. In: Advances in Neural Information Processing Systems 30, pp.~4765--4774. Curran Associates, Red Hook, NY, USA (2017)

\bibitem{wachter2017}
Wachter, S., Mittelstadt, B., Russell, C.: Counterfactual explanations without opening the black box: automated decisions and the GDPR. Harvard Journal of Law \& Technology \textbf{31}(2), 841--887 (2018)

\bibitem{mothilal2020}
Mothilal, R.K., Sharma, A., Tan, C.: Explaining machine learning classifiers through diverse counterfactual explanations. In: Proceedings of the 2020 conference on fairness, accountability, and transparency, pp.~607--617. ACM, New York, NY, USA (2020)

\bibitem{miller2019}
Miller, T.: Explanation in artificial intelligence: insights from the social sciences. Artificial intelligence \textbf{267}, 1--38 (2019)

\bibitem{rasmussen2021}
Rasmussen, M.H., Lefran\c{c}ois, M., Schneider, G.F., Pauwels, P.: BOT: the Building Topology Ontology of the W3C Linked Building Data Group. Semantic Web \textbf{12}(1), 143--161 (2021)

\bibitem{rosen2021}
Rosen, A.: Urban Mining Index: Entwicklung einer Systematik zur quantitativen Bewertung der Kreislaufkonsistenz von Baukonstruktionen in der Neubauplanung. Fraunhofer IRB Verlag, Stuttgart, Germany (2021)

\bibitem{lecue2020}
Lecue, F.: On the role of knowledge graphs in explainable AI. Semantic Web \textbf{11}(1), 41--51 (2020)

\bibitem{tiddi2022}
Tiddi, I., Schlobach, S.: Knowledge graphs as tools for explainable machine learning: a survey. Artificial Intelligence \textbf{302}, 103627 (2022)

\bibitem{bharadwaj2000}
Bharadwaj, A.S.: A resource-based perspective on information technology capability and firm performance: an empirical investigation. MIS quarterly \textbf{24}(1), 169--196 (2000)

\bibitem{melville2004}
Melville, N., Kraemer, K., Gurbaxani, V.: Review: information technology and organizational performance: an integrative model of IT business value. MIS quarterly \textbf{28}(2), 283--322 (2004)

\bibitem{shneiderman2022}
Shneiderman, B.: Human-Centered AI. Oxford University Press, Oxford, United Kingdom (2022)

\bibitem{rudin2019}
Rudin, C.: Stop explaining black box machine learning models for high-stakes decisions and use interpretable models instead. Nature machine intelligence \textbf{1}(5), 206--215 (2019)

\bibitem{jacovi2021}
Jacovi, A., Marasovi\'c, A., Miller, T., Goldberg, Y.: Formalizing trust in artificial intelligence: prerequisites, causes and goals of human trust in AI. In: ACM conference on fairness, accountability, and transparency, pp.~624--635. ACM, New York, NY, USA (2021)

\bibitem{lai2019}
Lai, V., Tan, C.: On human predictions with explanations and predictions of machine learning models: a case study on deception detection. In: Proceedings of the conference on fairness, accountability, and transparency, pp.~29--38. ACM, New York, NY, USA (2019)

\bibitem{lebo2013}
Lebo, T., Sahoo, S., McGuinness, D. (eds.): PROV-O: The PROV Ontology. W3C Recommendation, World Wide Web Consortium (2013). https://www.w3.org/TR/2013/REC-prov-o-20130430/, last accessed 2026/07/09

\bibitem{holten2018}
Rasmussen, M. H., Lefran\c{c}ois, M., Bondulel, M., Hviid, C. A., Karlsh\o{}j, J.: OPM: An ontology for describing properties that evolve over time. In: Proceedings of the 6th Linked Data in Architecture and Construction Workshop, pp. 24-33. CEUR-WS.org, Bonn, Germany (2018)

\end{thebibliography}
\end{document}